\documentclass[lettersize,journal]{IEEEtran}
\usepackage{amsmath,amsfonts}
\usepackage{algorithmic}
\usepackage{algorithm}
\usepackage{array}
\usepackage[caption=false,font=normalsize,labelfont=sf,textfont=sf]{subfig}
\usepackage{textcomp}
\usepackage{stfloats}
\usepackage{url}
\usepackage{verbatim}
\usepackage{graphicx}
\usepackage{cite}
\usepackage{flushend}
\usepackage{color}
\usepackage{booktabs}
\usepackage{multicol}
\usepackage{multirow}
\usepackage{caption}
\allowdisplaybreaks[4]  

\begin{document}

\title{\LARGE DeepOPF-U: A Unified Deep Neural Network to Solve AC Optimal Power Flow in Multiple Networks}

\author{Heng Liang, Changhong Zhao,~\IEEEmembership{Senior Member,~IEEE}
\thanks{This work was supported by Hong Kong Research Grants Council through grant ECS 24210220. Corresponding author: Changhong Zhao.}
\thanks{H. Liang and C. Zhao are with the Department of Information Engineering, the Chinese University of Hong Kong, New Territories, Hong Kong SAR, China
(e-mail:  lh021@ie.cuhk.edu.hk; chzhao@ie.cuhk.edu.hk).}
\thanks{This work has been submitted to the IEEE for possible publication. Copyright may be transferred without notice, after which this version may no longer be accessible.}
}


\maketitle

\begin{abstract}
The traditional machine learning models to solve optimal power flow (OPF) are mostly trained for a given power network and lack generalizability to today's power networks with varying topologies and growing plug-and-play distributed energy resources (DERs). 
In this paper, we propose DeepOPF-U, which uses \emph{one} unified deep neural network (DNN) to solve alternating-current (AC) OPF problems in different power networks, including a set of power networks that is successively expanding. 
Specifically, we design elastic input and output layers for the vectors of given loads and OPF solutions with varying lengths in different networks. The proposed method, using a single unified DNN, can deal with different and growing numbers of buses, lines, loads, and DERs. Simulations of IEEE $57/118/300$-bus test systems and a network growing from $73$ to $118$ buses verify the improved performance of DeepOPF-U compared to existing DNN-based solution methods.
\end{abstract}


\section{Introduction}

\IEEEPARstart{T}{here} have been increasing research efforts in leveraging the powerful approximation capabilities of deep neural networks (DNNs) to learn the high dimensional load-to-solution mappings of the important optimal power flow (OPF) problems. High-quality near-optimal OPF solutions can be instantly obtained from well trained DNN models, significantly accelerating the solution process compared to traditional solvers \cite{pan2023acopf,huang2022opfv}. However, most existing DNN models were built and trained only for a specific power network with a given topology. With the expansion of buses, lines, loads, and distributed energy resources (DERs) and the corresponding change of power network topologies, the existing DNN models need to be rebuilt through repeated trainings, incurring heavy storage and computation burdens.


Solving alternating-current (AC) OPF problems across multiple networks is challenging, due to the difference in network topologies, line admittances, lengths of load and solution vectors, et cetera.
Several methods have been developed recently to partly address the topology variation issue. 
For instance, \cite{topology2023liu} resampled the training data and retrained the DNNs in real time to adapt to the emerging new topologies, which is often computationally expensive. 
Reference \cite{convopf2023jia} integrated the topology labels into the training process, while \cite{zhou2023opfft} encoded discrete topology representations with line admittances into the DNN input. 
These methods, without retraining, can directly predict OPF solutions under flexible topologies. However, they are still limited to a fixed number of loads and generators and incapable of incorporating plug-and-play DERs in a network expansion setting. 
Reference \cite{decenopf2022Torre} learned a linear controller for an expanding radial network, which is not applicable to networks with general topologies.

In this paper, we propose DeepOPF-U, a novel approach that utilizes \emph{one} unified DNN to learn the load-to-solution mappings of AC OPF problems across multiple and expanding networks with different numbers of buses, lines, loads, and generators. 
The contribution of this work includes:
\begin{itemize} 
\item We design elastic DNN input and output layers with plug-and-play neurons, to adapt to the varying lengths of load and OPF solution vectors in different networks.
\item We design an incremental training process to sequentially update the weights of a unified DNN for multiple and expanding networks. The unified DNN can predict the OPF solutions in multiple networks without being retrained.
\end{itemize}
Simulations on IEEE $57/118/300$-bus test systems and a network growing from $73$ to $118$ buses demonstrate the adaptive mapping capability of the unified DNN without compromising solution quality. To the best of our knowledge, this is the first work to solve AC OPF problems across multiple networks using a single DNN.



\section{AC OPF across Multiple Power Networks}
Consider a series of power networks indexed by $k =1,...,K$. Denote network $k$ by $\{\mathcal{N}_k,\mathcal{E}_k\}$, where $\mathcal{N}_k$ and $\mathcal{E}_k$ collect the buses and lines, respectively. We aim to solve the following AC OPF problem in each network $k$:
\begin{subequations}\label{OPF}
\begin{alignat}{2} 
(\textbf{P}_k):&\min_{P^g_{i},Q^g_{i}} \sum_{i=1}^{\mathcal{N}_k^G} C_i(P^g_{i}) \\
\text{s.t.} &\sum_{j\in\mathcal{N}_k}\operatorname{Re}\{V_{i} V_{j}^*Y_{ij}^*\}=P_{i}^g-P_{i}^d, ~\forall i \in \mathcal{N}_k, \label{OPF:kiff:p}\\
 &\sum_{j\in\mathcal{N}_k}\operatorname{Im}\{V_{i} V_{j}^*Y_{ij}^*\}=Q_{i}^g-Q_{i}^d, ~\forall i \in \mathcal{N}_k, \label{OPF:kiff:q} \\
& \underline{P}^{g}_{i}\leq P^{g}_{i} \leq \overline{P}^{g}_{i}, ~\forall i \in \mathcal{N}_k^G, \label{OPF:limit:pg}\\
& \underline{Q}^{g}_{i}\leq Q^{g}_{i} \leq \overline{Q}^{g}_{i}, ~\forall i \in \mathcal{N}_k^G, \label{OPF:limit:qg}\\
& \underline{V}_i \leq |V_i| \leq \overline{V}_i,  ~\forall i \in \mathcal{N}_k,\label{OPF:limit:V}\\
& |V_i (V_i^*-V_j^*)Y_{ij}^*|\leq \overline{S}_{ij}, ~\forall (i,j) \in \mathcal{E}_k \label{OPF:limit:S}
\end{alignat}
\end{subequations}
where $\mathcal{N}_k^G$ denotes the set of buses with dispatchable generators. $P_{i}^g$ and $Q_{i}^g$, $P_{i}^d$ and $Q_{i}^d$ represent the active and reactive power generation, active and reactive power consumption at bus $i$, respectively. $V_i$ denotes the complex voltage at bus $i$. $Y_{ij}$ is the $(i,j)$-th entry of the network admittance matrix $\boldsymbol{Y}$. Constants $\underline{x}$ and $\overline{x}$ are the lower and upper limits of variable $x$. The generation cost at bus $i \in\mathcal{N}_k^G$ is $C_i(P_{i}^g)$.

Problem $\textbf{P}_k$ varies significantly across different networks $k$. Prior methods \cite{pan2023acopf,huang2022opfv,zhou2023opfft} solved OPF problems in a specific network by using a dedicated DNN. As a new network emerges or the network successively expands, a new DNN needs to be built and trained, which lacks generalizability. 

To overcome the limitation above, we design \emph{one} unified DNN to learn the load-to-solution mappings of AC OPF problems across multiple and expanding networks. The proposed approach, called DeepOPF-U, is applicable to: 
\begin{itemize}
    \item Multiple networks with different numbers of buses, lines, loads, DERs, and different topologies;
    \item An expanding network with increasing numbers of buses, lines, loads, and DERs. 
\end{itemize}

\section{Unified DNN for Multiple Power Networks}



\begin{figure}
    \centering
    \includegraphics[width=1.0\columnwidth]{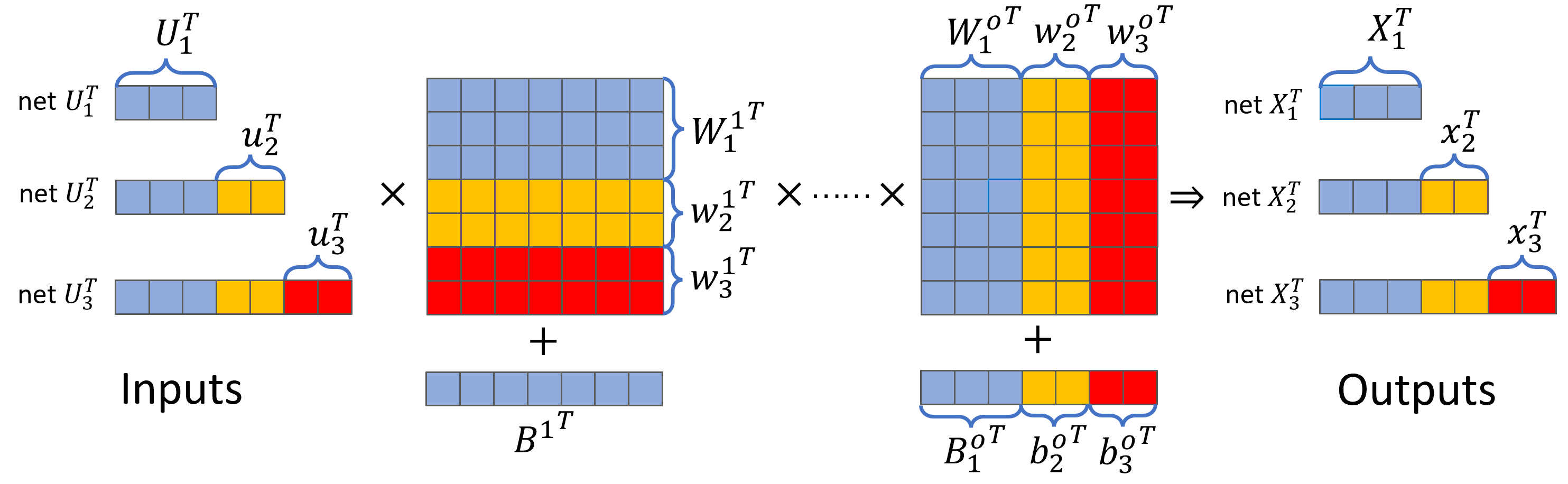}
    \caption{Demonstration of the unified DNN \eqref{DNN} (omitting activation functions). The sizes of parameter matrices $W^{1}_k$, $W^{o}_k$ and vector $B^o_k$ are elastic.}
    \label{fig:elasticnet}
\end{figure}


We design a single unified DNN to learn the load-to-solution mappings of AC OPF problems in multiple networks. For problems $\{\textbf{P}_k,~ k=1,...,K\}$, suppose their numbers of buses $N_1\leq\dots\leq N_K$. 
Let $U_1:=[{P^{d}_1}^{\top},{Q^{d}_1}^{\top}]^{\top}$ stack the active and reactive power loads of the smallest network $\{\mathcal{N}_1,\mathcal{E}_1\}$. Define $U_k:=[U_{k-1}^{\top},u_{k}^{\top}]^{\top}$, $W_{k}^1:=[W_{k-1}^1,w_k^1]$, $W^o_{k}:=[{W^o_{k-1}}^{\top},{w^o_k}^{\top}]^{\top}$, $B^o_{k}:=[{B^o_{k-1}}^{\top},{b^o_k}^{\top}]^{\top}$, $X_k:=[X_{k-1}^{\top},x_k^{\top}]^{\top}$ for $k=2,\dots,K$. The unified DNN is designed as a multi-layer fully-connected neural network:
\begin{subequations} \label{DNN}
\begin{alignat}{2}
U_k&=[U_{k-1}^{\top},u_k^{\top}]^{\top},\\
h^1&=\sigma([W^{1}_{k-1},w^{1}_{k}]U_k+B^1),\label{DNN:in}\\
h^i&=\sigma(W^ih^{i-1}+B^i), ~\forall i=2,...,L \\
\left[\begin{array}{c}
     X_{k-1}  \\
    x_{k}
\end{array}\right]&=\sigma^{\prime}\left(\left[\begin{array}{c}
     W^o_{k-1}  \\
    w^{o}_k
\end{array}\right]h^L+\left[\begin{array}{c}
     B^o_{k-1}  \\
    b^{o}_{k}
\end{array}\right]\right) \label{DNN:out}
\end{alignat}
\end{subequations}
where $u_k$ and $x_k$ are the increments of input and output vector lengths from network $(k-1)$ to $k$. Correspondingly, $w^{1}_k$, $w^{o}_k$ and $b^{o}_k$ are parameters of the plug-and-play neurons to be adaptively activated/deactivated according to the input and output lengths. 
ReLU function $\sigma(\cdot)$ and sigmoid 
function $\sigma^{\prime}(\cdot)$ are used as activation functions of the hidden layers and the output layer, respectively.
The elastic input and output layers \eqref{DNN:in}, \eqref{DNN:out} embed $K$ DNNs into a unified DNN in an incremental manner. 
The input, output, and parameters of the unified DNN are demonstrated in Figure \ref{fig:elasticnet}. 

The unified DNN is trained to minimize the following loss function corresponding to power network $k$:
\begin{alignat}{2}\label{loss_k} L_k:=\sum_{i\in\mathcal{N}_k}(|\hat{V}_i|-|V_i|)^2+ \gamma (\hat{\theta}_i-\theta_i)^2 \nonumber
\end{alignat}
where $|\hat{V}_i|$ and $\hat{\theta}_i$ are the voltage magnitude and phase angle at bus $i$ predicted by the DNN; $|V_i|$ and $\theta_i$ are their ground truth; factor $\gamma$ tunes the relative importance of the two terms.


Inspired by the once-for-all idea \cite{onceforall2020cai}, we design an incremental training strategy, which sequentially computes the losses $L_k$ and backpropagates it immediately after each $k$ to update the corresponding DNN parameters. 
Specifically, for each $k$,
the parameters for the input layer are updated as:
\begin{alignat}{2}
(\textbf{D}_k): W^{1}_k\leftarrow W^{1}_k-\alpha \nabla_{W^{1}_k} L_k \nonumber
\end{alignat}
where 
$\alpha>0$ is the learning rate; similarly for the update of output-layer parameters $W^o_{k}$ and $B^o_{k}$. 
After the complete training process $\textbf{D}_k$, $k=1,...,K$, the DNN can predict OPF solutions to $\text{\textbf{P}}_k$, $k =1,...,K$ without retraining.

\section{Numerical Experiments}


We consider two sets of test systems. The first set consists of the IEEE $57$-bus, $118$-bus, and $300$-bus feeders from PYPOWER. The second set consists of $73$-bus, $90$-bus, $106$-bus, and $118$-bus feeders, which (except the $118$-bus feeder itself) are all formed by removing buses (and loads and DERs on them) and lines from the IEEE $118$-bus feeder, to emulate a successively expanding case.

For the first set of feeders, load data are uniformly sampled within $[90\%,110\%]$ of the original loads. We sample  $150,000$ data points, $50,000$ for each feeder, among which $80\%$ form the training set and $20\%$ the test set.
For the second set of feeders, we use the 5-minute load profile of California ISO on May 20, 2020, which varies by up to $54\%$ from 00:00 to 23:59, to scale the original load at each bus. This serves as the base load and the test set. In each 5-minute slot, $400$ data points are sampled uniformly within $[90\%,110\%]$ of the base load, as the training set.
The conventional MIPS solver in PYPOWER is used to obtain the ground-truth OPF solutions at the sampled load data points.

\begin{table}
\caption{Details of unified DNNs built on PyTorch.}
\centering
\begin{tabular}{c|c|c|c}
\toprule\hline
 $\#$buses & Input layer  & Hidden layers & Output layer \\
\hline
57/118/300 & 84--374 & 1024/512/256 & 114--600   \\
73/90/106/118& 114--189 & 1024/512/256 & 146--236   \\
\hline
\bottomrule
\end{tabular}
\label{tab:config}
\end{table}

The unified DNNs are built on PyTorch, with details explained in Table \ref{tab:config}. The input and output layers have elastic lengths in the given ranges. 
The length of the output layer is twice the number of buses of the corresponding network, as both voltage magnitudes $|V_i|$ and phase angles $\theta_i$ are output. 
We apply the Adam optimizer with initial learning rate $\alpha=1\times10^{-3}$, mini-batch size $100$, and $500$ epochs. The learning rate halves every $50$ epochs. 

Following \cite{huang2022opfv,zhou2023opfft}, we use the metrics below to assess the performance. 1) \textit{Optimality loss}: the gaps between the OPF objective values predicted by DeepOPF-U and those returned by MIPS in PYPOWER. It is better to be closer to $0$. 2) \textit{Constraint satisfaction}: the percentage of inequality constraints \eqref{OPF:limit:pg}--\eqref{OPF:limit:S} being satisfied, including the active and reactive power generation limits $(\eta_{P^g}$ and $\eta_{Q^g})$, voltage magnitude limits $(\eta_{V})$, and branch flow magnitude limits $(\eta_{S_{\ell}})$. It is better to be closer to $100$. 3) \textit{Load satisfaction}: the average percentage of active ($\eta_{P^d}$) and reactive ($\eta_{Q^d}$) power loads being satisfied. It is better to be closer to $100$. 4) \textit{Speed-up}: the average number of times by which a DNN accelerates OPF solution compared to MIPS. The higher, the better.

\begin{table}
\setlength\tabcolsep{4pt}
\caption{Comparision between DeepOPF-U and DeepOPF-V \cite{huang2022opfv}.}
\centering
\begin{tabular}{c|c|c|c|c|c|c}
\toprule\hline 
\multirow{2}{*}{\text {Metric}} & \multicolumn{3}{c|}{\text {DeepOPF-U for all feeders}} & \multicolumn{3}{c}{\text {DeepOPF-V for each feeder}} \\
\cline { 2 - 7 } &\text{Case57}& \text {Case118} & \text{Case300} & \text {Case57}& \text{Case118}& \text{Case300}\\
\hline 
$\eta_{opt}(\%)$ & 0.44 & 0.35 & 0.60 & 0.31 & 0.32 & 0.43\\
$\eta_{V}(\%)$ & 100.0 & 100.0 & 100.0 & 100.0 & 100.0 & 100.0\\
$\eta_{P^g}(\%)$ & 100.0 & 99.64 & 99.80 & 100.0 & 99.64 & 99.80\\
$\eta_{Q^g}(\%)$ & 99.85 & 99.92 & 99.87 & 99.86 & 99.91 & 99.85\\
$\eta_{S_l}(\%)$ & 100.0 & 100.0 & 100.0 & 100.0 & 100.0 & 100.0\\
$\eta_{P^d}(\%)$ & 99.67 & 99.74 & 99.54 & 99.77 & 99.76 & 99.67\\
$\eta_{Q^d}(\%)$ & 99.63 & 99.70 & 99.47 & 99.67 & 99.75 & 99.63\\
speed-up &$\times$456 & $\times$487 & $\times$1299 &  $\times$459 & $\times$491 & $\times$1313\\
\hline
storage & \multicolumn{3}{c|}{5.5 MB} & \multicolumn{3}{c}{12.4 MB} \\
\hline\bottomrule
\end{tabular}
\label{tab:cmpsingle}
\end{table}

\subsection{Performance across Multiple Feeders}
\label{subsec:multiplenetwork}

We compare the proposed DeepOPF-U with DeepOPF-V \cite{huang2022opfv} in the first set of feeders to verify the capability of the former across multiple feeders. The results are shown in Table \ref{tab:cmpsingle}. 
The configurations of two methods are set the same for fair comparison. 
DeepOPF-U can predict AC OPF solutions in three feeders using a single DNN without compromising solution quality, while DeepOPF-V can only predict solutions in each single feeder with a separate DNN. 
As a result, DeepOPF-U uses $44\%$ the storage space of DeepOPF-V, which can be even smaller with more feeders.
Meanwhile, DeepOPF-U achieves a computation speed-up (compared to the conventional MIPS solver) similar to that of DeepOPF-V.

\begin{figure}
    \centering
    \includegraphics[width=1.0\columnwidth]{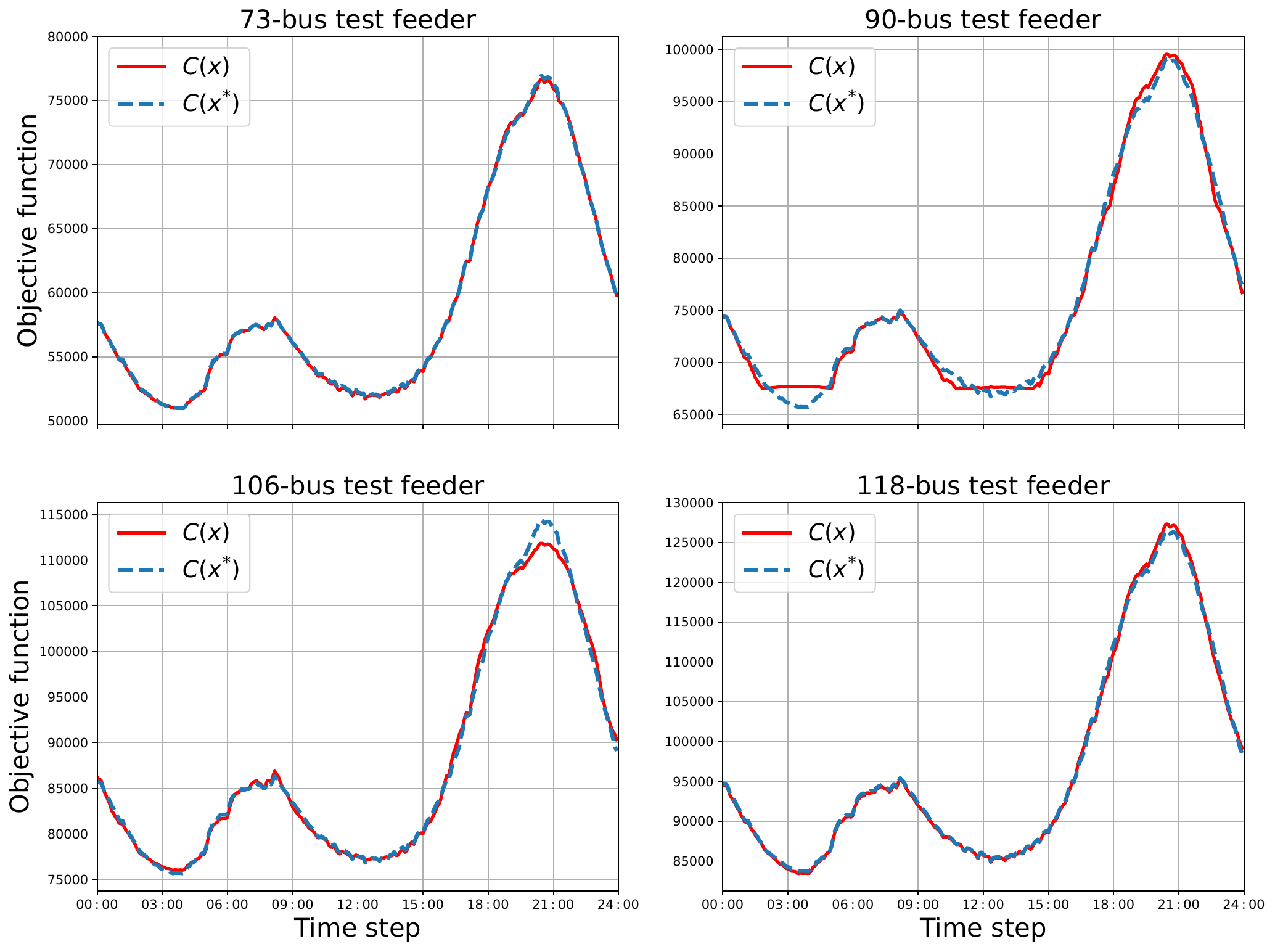}
    \caption{Time-varying OPF objective values in a set of successively expanding feeders. The red solid lines $C(x)$ show the values predicted by DeepOPF-U, and the blue dashed lines $C(x^*)$ are solved by MIPS.}
    \label{fig:tracking}
\end{figure}

\subsection{Performance in Expanding Feeders}

Consider the second set of feeders, which successively expands from $73$, $90$, $106$, to $118$ buses. 
Figure \ref{fig:tracking} shows their OPF objective values over time, solved by DeepOPF-U and the conventional MIPS solver. 
The result shows that DeepOPF-U can effectively track the OPF solutions in a time-varying (and load-varying) setting, with minor difference from the ground-truth solutions obtained by MIPS.

\begin{table}[!t]
\caption{Performance of three methods on IEEE $2,000$-bus feeder.}
\centering
\begin{tabular}{c|c|c|c}
\toprule\hline
Metric & DeepOPF-U  & DACOPF \cite{pan2023acopf} & EACOPF \cite{baker2020eacopf} \\
\hline
$\eta_{o p t}(\%)$ & 0.23  & 0.006 & 0.29\\
$\eta_{V}(\%)$ & 100.0  & 99.97 & 96.27\\
$\eta_{P^g}(\%)$ & 99.78  & 99.89& 99.88\\
$\eta_{Q^g}(\%)$ &  100.00 & 100.00 & 99.99 \\
$\eta_{S_l}(\%)$ & 99.58  & 99.60 & 99.47\\
$\eta_{P^d}(\%)$ & 99.82  & 100.0 & 100.0\\
$\eta_{Q^d}(\%)$ & 99.32  & 100.0 & 100.0\\
speed-up & $\times$4880  & $\times$184 & $\times$373\\
\hline
\bottomrule
\end{tabular}
\label{tab:sota}
\end{table}

\subsection{Comparision with State-of-the-Art Methods}

We compare the performance of DeepOPF-U with state-of-the-art DACOPF \cite{pan2023acopf} and EACOPF \cite{baker2020eacopf} (at their default settings in the papers) on the IEEE $2,000$-bus feeder, as shown in Table \ref{tab:sota}.  
Note that DeepOPF-U can solve OPF problems in $57$/$118$/$300$/$2000$-bus feeders with one unified NN, while the other two methods can only deal with one feeder with one NN. With such better generalizability, DeepOPF-U can still achieve good solution quality, especially more strict satisfaction of voltage constraints, and a significantly higher computation speed-up (compared to MIPS) than other two methods.


\section{Conclusion}

We proposed DeepOPF-U, a unified DNN  to learn the load-to-solution mappings of AC OPF problems across multiple and expanding power networks. 
Simulation results on IEEE $57/118/300$-bus test systems and a system expanding from $73$ to $118$ buses demonstrate the adaptive mapping capability, good solution quality, and satisfactory computational speed of the unified DNN. 

In the future, we will experiment on 
the best number of different power networks that can be effectively integrated into a unified DNN. We are also interested in exploring the similarities between OPF problems in different power networks, to characterize the performance of the unified DNN.


 





\end{document}